\documentclass[10pt,twocolumn,letterpaper]{article}

\usepackage[pagenumbers]{wacv}  %

\usepackage{booktabs,multirow,array}
\usepackage{graphicx}
\usepackage{xcolor}
\usepackage{tikz}
\usetikzlibrary{positioning,arrows.meta,fit,calc}
\usepackage{pgfplots}
\pgfplotsset{compat=1.17}

\definecolor{cCocap}{HTML}{D55E00}
\definecolor{cOurs}{HTML}{0072B2}
\definecolor{cTome}{HTML}{009E73}
\definecolor{cPool}{HTML}{E69F00}

\definecolor{wacvblue}{rgb}{0.21,0.49,0.74}
\usepackage[pagebackref,breaklinks,colorlinks,allcolors=wacvblue]{hyperref}

\title{Keyframes First, Motion Second: Revisiting Compressed-Domain\\Video Captioning under Limited Supervision}

\author{Asok B K\textsuperscript{*} \qquad Ashim Nepal\textsuperscript{*} \qquad Loknath Regmi\textsuperscript{\dag}\\[4pt]
{\small\textsuperscript{*}Equal contribution \quad \textsuperscript{\dag}Corresponding author}\\[2pt]
{\tt\small ashokbk215@gmail.com \quad official.nepalashim@gmail.com \quad lnregmi046@pcampus.edu.np}}

\begin{document}
\maketitle
\begin{abstract}
Compressed-domain video captioners such as CoCap read the H.264 bitstream directly: a pretrained encoder
reads the keyframes (I-frames), while motion and residual encoders, trained from scratch, turn the codec's
motion vectors and residuals into action tokens. This motion branch costs decoding time and supervision, both
scarce on edge devices. We ask when it pays off under limited supervision. In a controlled study on 2,500
VATEX training clips, we fix a frozen SigLIP~2 keyframe encoder, a GPT-2 caption decoder and the training
schedule, vary only the visual tokens the decoder receives (34 configurations), and report held-out test
CIDEr with paired bootstrap intervals and three seeds. Motion encoders trained from scratch bring no
measurable benefit: CoCap-style action tokens change CIDEr by $-0.55$ (95\% CI $[-2.3, +1.2]$), and six other
uses of motion do not help. Warm-started from a model trained on $48\times$ the caption samples, they add $+1.3$
CIDEr on average, though not significantly across seeds, and only while the keyframe tokens stay frozen.
Reading the keyframes better matters more: fine-tuning SigLIP~2's own attention-pooling head adds 5.4 CIDEr,
and one to eight such tokens beat the 16 CoCap-style tokens on every seed while decoding only I-frames
(7.3\,ms instead of \mbox{44--114}\,ms per clip). The same holds on MSVD's full split in CoCap's own video
encoding, where the readout leads by 14.2 CIDEr and action tokens bring no reliable gain, also with clean
single-reference motion vectors. Under limited
supervision, keyframes come first and motion second.
\end{abstract}

\section{Introduction}\label{sec:intro}

\begin{figure*}[t]
\centering
\begin{tikzpicture}[
  font=\footnotesize,
  box/.style={draw=black!55, rounded corners=2pt, align=center, minimum height=8.5mm, inner sep=3pt,
              text width=26mm, fill=white},
  frozen/.style={box, fill=black!6},
  tuned/.style={box, fill=cOurs!14},
  scratch/.style={box, fill=cCocap!14},
  plain/.style={box, draw=none, fill=none},
  arr/.style={-{Stealth[length=1.8mm]}, semithick, draw=black!65},
  count/.style={font=\scriptsize, text=black!70, midway, above},
  lane/.style={font=\footnotesize\bfseries, anchor=west, inner sep=0pt},
]
\node[lane] at (-0.2, 0.2) {(a) CoCap-style tokens (re-implemented in the same pipeline)};
\node[plain, text width=14mm] (abs) at (0.6,-1.25) {Video\\bitstream};
\node[frozen] (ai) at (3.3,-0.65) {I-frames\\(decoded)};
\node[box] (ap) at (3.3,-1.85) {P-frames: motion\\vectors, residuals};
\node[frozen] (aie) at (6.7,-0.65) {SigLIP~2\\pooled token};
\node[scratch, text width=30mm] (ape) at (6.7,-1.85) {Motion, residual and\\action encoders};
\node[box, text width=18mm] (ait) at (9.75,-0.65) {8 context\\tokens};
\node[box, text width=18mm] (apt) at (9.75,-1.85) {8 action\\tokens};
\node[tuned, text width=16mm] (adec) at (12.55,-1.25) {GPT-2\\decoder};
\node[plain, text width=20mm] (aout) at (15.2,-1.25) {\textbf{16 tokens}\\test CIDEr 28.7};
\draw[arr] (abs.east) -- (ai.west);
\draw[arr] (abs.east) -- (ap.west);
\draw[arr] (ai) -- (aie);  \draw[arr] (ap) -- (ape);
\draw[arr] (aie) -- (ait); \draw[arr] (ape) -- (apt);
\draw[arr] (ait.east) -- (adec.west);
\draw[arr] (apt.east) -- (adec.west);
\draw[arr] (adec) -- (aout);
\node[lane] at (-0.2,-3.05) {(b) Keyframe-only readout (this study)};
\node[plain, text width=14mm] (bbs) at (0.6,-4.05) {Video\\bitstream};
\node[frozen] (bi) at (3.3,-4.05) {I-frames only\\(21\% of coded bytes)};
\node[frozen, text width=22mm] (be) at (6.5,-4.05) {SigLIP~2 patches\\$8\times196$};
\node[tuned, text width=20mm] (br) at (9.95,-4.05) {Pooling head\\(pretrained start)};
\node[tuned, text width=16mm] (bdec) at (12.55,-4.05) {GPT-2\\decoder};
\node[plain, text width=20mm] (bout) at (15.2,-4.05) {\textbf{1 token}\\test CIDEr 34.1};
\draw[arr] (bbs) -- (bi); \draw[arr] (bi) -- (be);
\draw[arr] (be) -- node[count] {1,568} (br);
\draw[arr] (br) -- node[count] {1} (bdec);
\draw[arr] (bdec) -- (bout);
\node[frozen, text width=6mm, minimum height=3mm, inner sep=1pt] (k1) at (2.4,-5.15) {};
\node[anchor=west, font=\scriptsize] at (k1.east) {frozen, pretrained};
\node[tuned, text width=6mm, minimum height=3mm, inner sep=1pt] (k2) at (6.0,-5.15) {};
\node[anchor=west, font=\scriptsize] at (k2.east) {trained from a pretrained start};
\node[scratch, text width=6mm, minimum height=3mm, inner sep=1pt] (k3) at (11.2,-5.15) {};
\node[anchor=west, font=\scriptsize] at (k3.east) {trained from scratch};
\end{tikzpicture}
\caption{The two decoder inputs compared throughout. (a) CoCap-style tokens: per GOP, one context token
(SigLIP~2's pooled I-frame embedding) and one action token from motion vectors and residuals, whose motion,
residual and action encoders train from scratch. (b) Keyframe-only readout: only I-frames are decoded, and SigLIP~2's own
attention-pooling head, fine-tuned from its pretrained weights, reads all 1,568 patch tokens into one decoder
token. Encoder, decoder, data and schedule are identical in (a) and (b); test CIDEr on 1,000 held-out clips.}
\label{fig:pipeline}
\end{figure*}
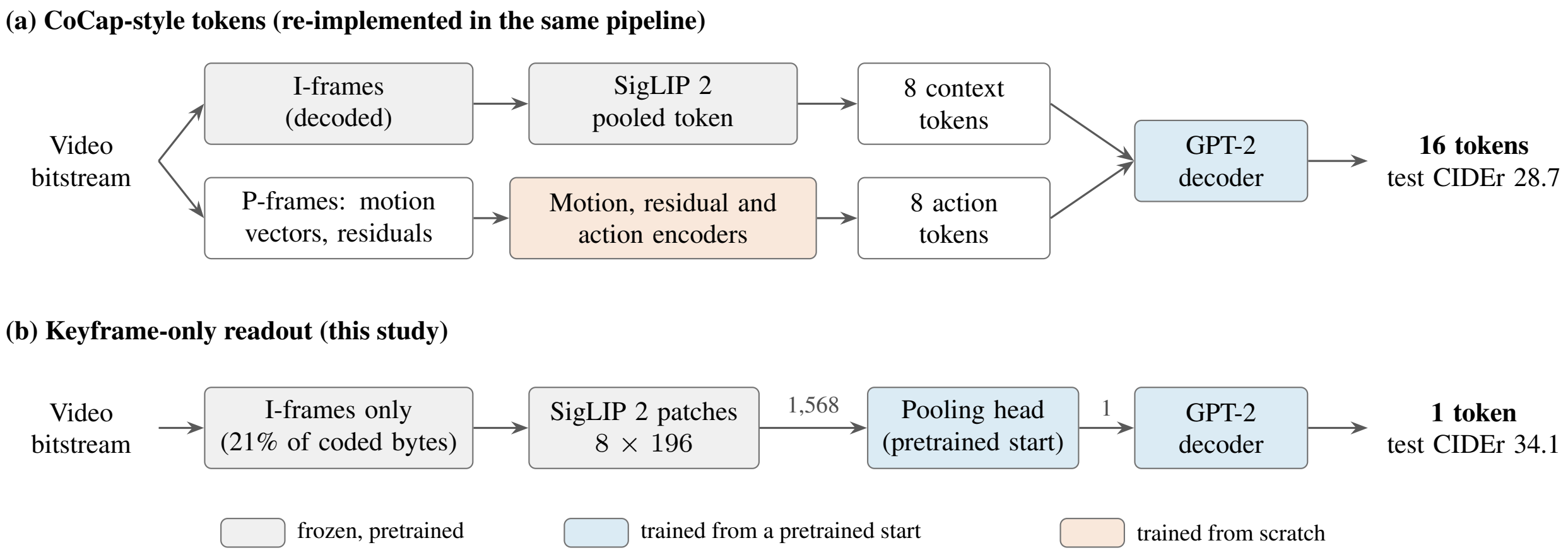

Cameras on drones, in surveillance systems and on other edge devices deliver compressed H.264 bitstreams,
not frames. A compressed video already separates appearance from change. Each group of pictures (GOP)
starts with an intra-coded I-frame, and the following P-frames store motion vectors and residuals relative
to earlier frames. Compressed-domain methods use this structure to avoid decoding and sampling every
frame~\cite{wu2018coviar,shen2023cocap}. For captioning, CoCap~\cite{shen2023cocap} encodes each I-frame into a
context feature with a pretrained I-frame encoder. Lightweight motion and residual encoders and an action
encoder turn the motion vectors and residuals of the same GOP into an action feature, and a multimodal
decoder reads both for eight GOPs. With full training data on MSR-VTT, this motion branch adds $+2.1$ CIDEr
over I-frames alone~\cite{shen2023cocap}.

That gain may depend on how well the motion encoders are trained. They learn from scratch from captions,
whereas the I-frame encoder inherits large-scale image-text pretraining, and image encoders such as
SigLIP~2~\cite{tschannen2025siglip2} have since become much stronger. Edge deployments also rarely come with
the full caption corpora that CoCap trains on. A prior study~\cite{nepal2026substitution} replaced CoCap's
I-frame encoder with SigLIP~2 and its decoder with GPT-2~\cite{radford2019gpt2}, but kept CoCap's 16 tokens.
Two questions remain open. What should the decoder receive from the keyframes? And does codec motion still
pay off when supervision is limited?

We answer both under strict control. The I-frame encoder (SigLIP~2), the decoder (GPT-2), the training clips
(2,500 from VATEX~\cite{wang2019vatex}), the schedule and the decoding are fixed. Only the visual tokens vary,
from 1 to 1,568, with and without codec motion (Fig.~\ref{fig:pipeline}). Every comparison is paired over the
same clips, the main claims are repeated over three seeds, and all numbers come from 1,000 held-out test clips
that played no part in any design decision. To separate the value of motion from how its encoders are
trained, we also warm-start the motion encoders from a CoCap model trained on $48\times$ our caption samples,
and add the action tokens to fine-tuned context tokens. Finally, we repeat the key configurations on
MSVD~\cite{chen2011msvd}, using its full official split and CoCap's own video encoding.

\paragraph{Contributions.}
\begin{itemize}
\item \textbf{Codec motion under limited supervision.} With motion encoders trained from scratch, CoCap-style
  action tokens change test CIDEr by $-0.55$ (95\% CI $[-2.3, +1.2]$). Six other uses of motion do not help
  either, including on action verbs and on captions that describe direction. Warm-started encoders add $+1.3$
  CIDEr on average, but not significantly across seeds, and not next to fine-tuned context tokens. On MSVD in CoCap's own encoding, action tokens bring no reliable gain, also with clean single-reference
  motion vectors (Secs.~\ref{sec:motion}--\ref{sec:msvd}).
\item \textbf{Reading the keyframes.} Fine-tuning the I-frame encoder's own attention-pooling head from its
  pretrained weights adds 5.4 CIDEr over three seeds, and \mbox{4--5.5} CIDEr over a random start. More keyframes also
  help ($+5.5$ CIDEr from one to eight), so the clips are not single-frame problems (Sec.~\ref{sec:readout}).
\item \textbf{A small readout suffices.} One fine-tuned readout token is as accurate as eight on VATEX, and it
  beats the 16 CoCap-style tokens on every seed of both datasets, even when their motion encoders are
  warm-started.
  The lead is 8.8 CIDEr on VATEX after 12 epochs and 14.2 on MSVD (Secs.~\ref{sec:msvd},~\ref{sec:budget}).
\item \textbf{Decode cost.} The readout needs only I-frames. They decode in 7.3\,ms per clip, against \mbox{44--114}\,ms for the tested readers that also return motion vectors and residuals, which saves \mbox{19--36\%} of the time per clip
  (Sec.~\ref{sec:efficiency}).
\end{itemize}

Our setting mirrors on-device use: the input is the H.264 stream itself, labeled data are scarce, and
decoding and token budgets are tight. We did not, however, test aerial footage or edge hardware. This is a
controlled study, not a benchmark entry. CoCap reports 52.7 test CIDEr on VATEX with its full training set and
a ViT-B/16 encoder~\cite{shen2023cocap}. Our models see a tenth of its training clips and one caption per clip
and epoch, and are scored on a 1,000-clip subset of the test split, so absolute scores are not comparable.
``CoCap-style'' denotes CoCap's token design re-implemented with the same encoder, decoder, data and schedule
as every other configuration.

\section{Related Work}\label{sec:related}

\paragraph{Compressed-domain video.}
CoViAR~\cite{wu2018coviar} and DMC-Net~\cite{shou2019dmcnet} recognized actions directly from I-frames,
motion vectors and residuals, and earlier work built motion descriptors from codec
vectors~\cite{kantorov2014mpeg}. CoCap~\cite{shen2023cocap}
brought the decomposition to captioning with an end-to-end transformer: an I-frame encoder initialized from
CLIP, motion and residual encoders, an action encoder that fuses them with the I-frame context, and a
multimodal decoder. CVT5~\cite{pirhadi2025cvt5} captions compressed video densely and found residuals
harmful on SoccerNet. Video-LaVIT~\cite{jin2024videolavit} and EMA~\cite{zhao2025ema} build video language models on keyframes and
motion vectors; MeToM~\cite{metom2026} and CodecSight~\cite{codecsight2026} use residuals, packet sizes and
motion magnitude to decide where such models spend tokens; and
CoPE-VideoLM~\cite{cope2026} replaces P-frame encoding with motion and residual tokens whose encoders are
pretrained at scale to align with the image encoder. These works use codec signals for efficiency, or train motion encoders with large-scale alignment. We
instead measure what codec motion adds to accuracy when its encoders learn from limited caption supervision,
and when they start from a better-trained model.

\paragraph{Captioning with pretrained models.}
End-to-end video captioners such as SwinBERT~\cite{lin2022swinbert} learn from frames; prefix
decoders~\cite{mokady2021clipcap,ramos2023smallcap} connect frozen image-text encoders to pretrained
language models. VideoCoCa~\cite{yan2022videococa} reuses CoCa's~\cite{yu2022coca} pretrained attentional
poolers on the flattened tokens of all frames, and pretrained image-text models adapt well to video
captioning~\cite{secretlycaptioners2025}. Single-frame models are competitive on many video-language
benchmarks~\cite{lei2023singleframe,buch2022atp}, and our keyframe ablation separates that effect from the
value of codec motion.

\paragraph{Token reduction and readouts.}
ToMe~\cite{bolya2023tome} merges similar tokens by bipartite matching; TokenLearner~\cite{ryoo2021tokenlearner}
learns attention maps that produce a few tokens per frame; video methods merge static or redundant tokens
over time~\cite{choudhury2024rlt,sttm2025,tao2024dycoke}. Learned queries that cross-attend to all visual
tokens are the standard connector~\cite{jaegle2021perceiver,alayrac2022flamingo,li2023blip2}. We compare
these families as interfaces between a frozen keyframe encoder and a caption decoder, and study the
attention-pooling head that SigLIP~2 itself uses to form its image embedding.

\section{Method}\label{sec:method}

\subsection{Compressed Video and Captioner}\label{sec:setup}

\paragraph{Compressed video.}
VATEX clips (about 10\,s) are re-encoded with H.264 at $224\times224$, with an I-frame every 30 frames, no
B-frames and up to three reference frames per P-frame (x264's default, which CoCap's own conversion script
also keeps). This gives 9.7 GOPs per clip on
average. The cache does not store which reference frame each block uses, so motion vectors that point to different references are mixed, which may weaken the motion signal; Sec.~\ref{sec:msvd} tests single-reference vectors. We sample
eight GOPs uniformly. The I-frame encoder, SigLIP~2 base/16~\cite{tschannen2025siglip2}, turns each I-frame
into 196 patch tokens (1,568 in total), cached once with the encoder frozen. Each P-frame provides motion
vectors on the codec's quarter-resolution grid ($56\times56\times2$) and a residual ($56\times56\times3$).

\paragraph{Caption decoder.}
GPT-2 small (124M)~\cite{radford2019gpt2} serves as a prefix decoder~\cite{mokady2021clipcap}: visual tokens
pass a LayerNorm and a linear projection and precede the caption. Visual tokens take no GPT-2 positions.
Instead, each receives a Fourier embedding of its source $(t, y, x)$ coordinates through a zero-initialized
projection, so every token budget uses the same text positions. The tied token embedding stays frozen, and
the twelve transformer blocks train.

\subsection{Decoder Inputs Compared}\label{sec:inputs}

\paragraph{CoCap-style tokens.}
Following CoCap~\cite{shen2023cocap}, each GOP gives one context token and one action token, 16 tokens for
eight GOPs. The context token is SigLIP~2's image embedding of the I-frame. The action token comes from motion
and residual encoders and an action encoder that fuses them with the I-frame's patch tokens. CoCap fine-tunes
its I-frame encoder end to end. Here, as in every configuration, the SigLIP~2 backbone stays frozen, so the
context token is a frozen image embedding. The per-I-frame readout below is the same token with the pooling
head fine-tuned, and Sec.~\ref{sec:warm} also adds the action tokens to it. The motion modules either train
from scratch or are warm-started from a CoCap model trained end to end, with CoCap's multimodal decoder on the
same SigLIP~2 features, using all ten captions of about 5,000 VATEX training clips for 12 epochs.

\paragraph{Merging and pooling.}
Average pooling over $1\times2\times2$ patch blocks, and ToMe~\cite{bolya2023tome} bipartite merging over all
1,568 patches with size-weighted averaging, in one shot or gradually. ToMe-$N$ denotes $N$ output tokens.

\paragraph{Attention-pooling readouts.}
SigLIP~2 has no class token. Its image embedding comes from a multihead attention-pooling (MAP) head, in which
a learned probe query attends to the final patch tokens. Applied to a frame's cached patches, the head
reproduces the cached image embedding exactly (cosine 1.0000), so applied to any subset of patches it embeds
that subset. We use it as a readout in three forms: the \emph{per-I-frame readout} (one token per I-frame),
the \emph{ToMe-cluster readout} (one token per ToMe cluster) and the \emph{$K$-query readout}, in which
$K\in\{1, 8, 16, 32\}$ queries start at the pretrained probe and attend to all 1,568 patches. The head is
frozen, fine-tuned from its pretrained weights, or re-initialized at random and trained.

\paragraph{Other uses of motion.}
We also test ToMe variants whose merge scores are lowered for patches with high motion-vector or residual
energy, or for patches that CoCap's action tokens attend to. Other variants attach seven motion statistics per
patch to every token through a zero-initialized projection, or give the decoder eight merged tokens plus eight
action tokens.

\subsection{Training and Evaluation}\label{sec:protocol}

\paragraph{Data.}
Training uses 2,500 VATEX~\cite{wang2019vatex} clips, with one of their ten English captions chosen at random
per clip and epoch, for five epochs (12,500 caption samples). Development decisions use 500 validation clips,
and final numbers use 1,000 held-out clips from VATEX's public test split. No source video or caption is
shared between splits. MSVD~\cite{chen2011msvd} uses its full official split and CoCap's own encoding
(Sec.~\ref{sec:msvd}).

\paragraph{Training.}
We train with batch 1 and gradient accumulation 8, a learning rate of $10^{-4}$ for new modules and
$5\times10^{-5}$ for GPT-2, 10\% warm-up and decay 0.8 per epoch, in bf16 with label-smoothed cross-entropy,
on one 4\,GB laptop GPU. GPT-2 stays frozen during the first epoch while the visual side trains, and the
sequence starts with a BOS token (Sec.~\ref{sec:regimes}). Warm-started motion modules train at a quarter of
the learning rate. Four key configurations are also trained for 12 epochs.

\paragraph{Evaluation.}
We decode greedily and report CIDEr-D~\cite{vedantam2015cider}, BLEU-4, METEOR and ROUGE-L. Differences
between configurations are paired bootstrap estimates over clips (5,000
resamples)~\cite{koehn2004bootstrap}, which capture clip sampling. Seeds 42, 43 and 44 capture training randomness for the configurations behind the main claims, and for those
we also report the mean and standard deviation over seeds. Where a difference is pooled over seeds, each clip's
CIDEr difference is first averaged over the seeds and the clips are then resampled; such intervals still cover
clip sampling only, so the per-seed differences are given alongside. Training uses non-deterministic GPU kernels, so a seed fixes
data order, caption sampling and initialization but not the exact training trajectory, and the spread over
seeds includes this variation.

\section{Experiments}\label{sec:experiments}

All numbers are CIDEr $\times100$ on the 1,000 held-out test clips with seed 42, unless stated otherwise.
Test scores use the checkpoint with the highest validation CIDEr, which is the last epoch for 52 of 59 runs,
and validation scores are taken at the last epoch. $\Delta$ denotes a paired difference with its 95\% bootstrap
interval, and the best value in each table column is set in bold.

\subsection{Main Results}\label{sec:main}

\begin{table*}[t]
\centering
\small
\caption{Main comparison on VATEX (2,500 training clips). All configurations read the same eight I-frames;
only the CoCap-style tokens also read motion vectors and residuals. Mean $\pm$ standard deviation over the
three seeds. With seed 43 the CoCap-style configuration fell into the high-loss training regime, and with
seed 44 ToMe-392 did (Sec.~\ref{sec:regimes}).}
\label{tab:main}
\begin{tabular}{@{}l c c c c c c c c@{}}
\toprule
 & & \multicolumn{4}{c}{Test CIDEr} & \multicolumn{3}{c}{Test, seed 42} \\
\cmidrule(lr){3-6}\cmidrule(lr){7-9}
Decoder input & Tokens & Seed 42 & Seed 43 & Seed 44 & Mean $\pm$ std & BLEU-4 & METEOR & ROUGE-L \\
\midrule
CoCap-style (8 context + 8 action) & 16 & 28.66 & 20.72 & 26.52 & 25.30 $\pm$ 4.11 & 21.88 & 19.76 & 44.20 \\
Context tokens only, no action tokens & 8 & 29.21 & 28.52 & 27.45 & 28.39 $\pm$ 0.89 & 21.77 & 19.59 & 43.87 \\
Average pooling & 392 & 22.25 & -- & -- & -- & 20.34 & 18.29 & 42.57 \\
ToMe & 16 & 30.39 & -- & -- & -- & 23.42 & 19.67 & 44.88 \\
ToMe & 392 & 36.75 & 35.51 & 25.35 & 32.54 $\pm$ 6.25 & \textbf{25.96} & 20.70 & \textbf{45.93} \\
\midrule
Per-I-frame readout, fine-tuned & 8 & 35.44 & 32.15 & 33.74 & 33.78 $\pm$ 1.64 & 23.83 & 20.59 & 45.14 \\
One-query readout, fine-tuned & 1 & 34.13 & 34.06 & 33.27 & 33.82 $\pm$ 0.48 & 23.29 & 20.26 & 44.88 \\
16-query readout, fine-tuned & 16 & 36.14 & \textbf{36.34} & \textbf{35.52} & \textbf{36.00 $\pm$ 0.43} & 23.62 & 20.62 & 45.17 \\
ToMe-cluster readout, fine-tuned & 16 & \textbf{37.25} & -- & -- & -- & 24.18 & \textbf{20.80} & 45.25 \\
\bottomrule
\end{tabular}
\end{table*}

Table~\ref{tab:main} summarizes the comparison; Table~S1 in the supplementary material lists all 34
configurations. Three
regularities hold on both splits and across seeds. Merging keeps far more than pooling at equal budget. A
fine-tuned readout of the keyframe patches into one to sixteen tokens outperforms the 16 CoCap-style tokens.
Adding action tokens to the context tokens does not help. The fine-tuned readout over ToMe clusters is the
most accurate configuration on test (37.25). Test scores are \mbox{8--12.5} CIDEr below validation, and neighboring
configurations sometimes swap places between the two splits.

\subsection{Does Codec Motion Help?}\label{sec:motion}

The cleanest test removes CoCap's action tokens and keeps its context tokens (Table~\ref{tab:motion}, first
row). Adding the action tokens changes test CIDEr by $-0.55$ (CI $[-2.3, +1.2]$), and by $-1.41$
(CI $[-3.3, +0.5]$) after 12 epochs. For this seed, read as one-sided bounds, they exclude a benefit larger than 1.2 CIDEr (0.5 after 12
epochs); smaller benefits remain possible. For scale, CoCap attributes $+2.1$ CIDEr (3.9\%
relative) to its motion branch on MSR-VTT with full training data~\cite{shen2023cocap}. At our CIDEr level the
same relative gain would be about $+1.1$, inside the five-epoch interval but above the 12-epoch bound. Over
three seeds the action tokens never help (Table~\ref{tab:warm}). With seed 43 they also pushed training into
the high-loss regime (Sec.~\ref{sec:regimes}), and without that seed the average change is $-0.74$. Our VATEX motion vectors mix displacements to up to three reference frames (Sec.~\ref{sec:setup}); on MSVD, single-reference vectors give the CoCap-style tokens no statistically resolved improvement
(Sec.~\ref{sec:msvd}). This test also uses frozen context tokens
and motion encoders trained from scratch. Sec.~\ref{sec:warm} changes both.

The other rows of Table~\ref{tab:motion} test specific interfaces for motion: motion or residual energy that
decides which patches merge, motion statistics attached to every token, and action tokens placed next to
merged patches. None helps, and five of the six are negative with intervals that exclude or touch zero.
Because CIDEr weights nouns heavily, we also checked action verbs. We take the first \emph{-ing} word after
\emph{is/are} as a caption's main verb and count a hit when it matches a main verb of the clip's ten
references. A second probe restricts CIDEr to the 257 validation clips whose references use direction or
directed-motion words in at least three of ten captions. No form of motion shows a resolved gain on either probe
(every point estimate is at or below zero), while more keyframes improve both (last row). 

\begin{table*}[t]
\centering
\small
\caption{Codec motion with motion encoders trained from scratch (seed 42). Each row adds motion to an
otherwise identical configuration; $\Delta$ = with motion $-$ without, 95\% bootstrap intervals. Guided
merging is compared with the ToMe variant it modifies (one-shot ToMe-16, gradual ToMe-392). The probes use
validation captions: main-verb hit rate (points) and CIDEr on the 257 clips whose references describe
direction. The last row is a control that adds keyframes instead of motion. Single seed, so indicative: the same
configuration can vary by \mbox{1--8} CIDEr across seeds.}
\label{tab:motion}
\begin{tabular}{@{}l c c c c@{}}
\toprule
 & \multicolumn{2}{c}{$\Delta$ CIDEr} & \multicolumn{2}{c}{Validation probes} \\
\cmidrule(lr){2-3}\cmidrule(lr){4-5}
Comparison & Validation & Test & $\Delta$ verb hit & $\Delta$ CIDEr, direction \\
\midrule
Context tokens $\pm$ 8 action tokens & $-0.58$ [$-3.4$, $+2.2$] & $-0.55$ [$-2.3$, $+1.2$] & $-1.0$ [$-4.6$, $+2.4$] & $-1.4$ [$-5.2$, $+2.5$] \\
ToMe-16 vs.\ 8 ToMe + 8 action & $-4.51$ [$-7.4$, $-1.7$] & $-1.99$ [$-3.8$, $-0.2$] & $-0.6$ [$-4.6$, $+3.6$] & $-5.6$ [$-9.3$, $-1.8$] \\
Motion-guided ToMe-16 & $-4.89$ [$-7.6$, $-2.1$] & $-2.93$ [$-4.7$, $-1.1$] & $-1.6$ [$-4.8$, $+1.6$] & $-5.8$ [$-9.5$, $-2.2$] \\
Motion-guided gradual ToMe-392 & $-1.42$ [$-3.8$, $+1.0$] & $-1.83$ [$-3.4$, $-0.2$] & $-2.4$ [$-5.5$, $+0.6$] & $-0.1$ [$-3.4$, $+3.4$] \\
Residual-guided gradual ToMe-392 & $-1.65$ [$-4.0$, $+0.8$] & $-1.76$ [$-3.5$, $-0.2$] & $+0.0$ [$-2.8$, $+2.8$] & $-0.6$ [$-3.5$, $+2.3$] \\
Motion statistics on tokens & $-1.07$ [$-3.9$, $+1.8$] & $-1.61$ [$-3.2$, $+0.0$] & $-4.4$ [$-7.9$, $-1.2$] & $-2.2$ [$-6.4$, $+2.0$] \\
\quad confidence-weighted & $-1.97$ [$-4.6$, $+0.7$] & $-0.29$ [$-1.9$, $+1.3$] & -- & -- \\
\midrule
Control: 1 $\to$ 8 keyframes & $\mathbf{+6.39}$ [$+3.4$, $+9.6$] & $\mathbf{+5.47}$ [$+3.5$, $+7.4$] & $\mathbf{+7.7}$ [$+3.8$, $+11.3$] & $\mathbf{+7.0}$ [$+2.9$, $+11.1$] \\
\bottomrule
\end{tabular}
\end{table*}

\subsection{Fine-Tuned Context, Warm-Started Motion}\label{sec:warm}

The test above uses frozen context tokens and motion modules trained from scratch. Table~\ref{tab:warm}
changes both. First, it adds the action tokens to fine-tuned context tokens, which is CoCap's 16-token layout
with the per-I-frame readout of Sec.~\ref{sec:readout}. Second, it warm-starts the motion, residual and action
encoders from a CoCap model trained on about 5,000 VATEX training clips, ours among them, with all ten
captions for 12 epochs. That is twice our clips and $48\times$ our caption samples, with no test clips. The
warm-started encoders therefore also carry supervision from 2,500 extra clips, so any gain is an upper bound on
what better-trained motion adds. The source model was selected on a validation split that contains our
development clips, so Table~\ref{tab:warm} reports test CIDEr only.

Warm-starting moves the CoCap-style design in the expected direction. On top of frozen context tokens,
warm-started action tokens add $+1.27$ CIDEr on average, against $-3.09$ from scratch, and no seed collapses.
The bootstrap interval of this gain, $[-0.0, +2.6]$, covers clip sampling only. Across seeds the gain is not
significant: one seed is negative, and a $t$-interval over the three seed means is $[-3.8, +6.3]$. Its size matches CoCap's relative full-data gain (about $+1.1$ at our level).

Next to fine-tuned context tokens, the result reverses. The action tokens lower CIDEr whether trained from
scratch ($-2.16$) or warm-started ($-4.21$), so in our pipeline the motion gain did not add to the larger gain
from fine-tuning the readout ($+5.38$, CI $[+4.2, +6.6]$). The action encoder attends to the same frozen
I-frame patches in both cases, so its input does not change. With action tokens present, however, fine-tuning
the readout no longer helps (29.57 vs.\ 29.67 averaged over seeds). The decoder seems to rely on the action
tokens' view of the I-frame instead of the readout, but we did not test this further. Even with warm-started
motion encoders, the 16 CoCap-style tokens trail the one-token readout by 4.15 CIDEr (CI $[+2.8, +5.5]$).

\begin{table}[t]
\centering
\small
\caption{Adding CoCap's eight action tokens to frozen or fine-tuned context tokens, with motion modules
trained from scratch or warm-started from a CoCap model trained on $48\times$ the caption samples (test
CIDEr). $\Delta$: change from the context-token row above, averaged over seeds, with its 95\% bootstrap
interval over clips.}
\label{tab:warm}
\setlength{\tabcolsep}{3.5pt}
\begin{tabular}{@{}l c c c l@{}}
\toprule
 & \multicolumn{3}{c}{Seed} & \\
\cmidrule(lr){2-4}
Decoder input & 42 & 43 & 44 & $\Delta$ \\
\midrule
Frozen context tokens & 29.21 & 28.52 & 27.45 & \\
\quad + action, scratch & 28.66 & 20.72 & 26.52 & $-3.09$ {\scriptsize$[-4.1, -2.1]$} \\
\quad + action, warm & 28.23 & 30.36 & 30.41 & $\mathbf{+1.27}$ {\scriptsize$[-0.0, +2.6]$} \\
\midrule
Fine-tuned context & \textbf{35.44} & \textbf{32.15} & 33.74 & \\
\quad + action, scratch & 28.21 & 29.82 & \textbf{36.83} & $-2.16$ {\scriptsize$[-3.3, -1.0]$} \\
\quad + action, warm & 28.83 & 29.22 & 30.66 & $-4.21$ {\scriptsize$[-5.6, -2.9]$} \\
\bottomrule
\end{tabular}
\end{table}

\subsection{MSVD in CoCap's Own Encoding}\label{sec:msvd}

To test whether these results depend on our encoding or on VATEX, we repeat the four key configurations on
MSVD~\cite{chen2011msvd} in CoCap's own encoding: H.264 with an I-frame every 60 frames, short side 240, no
B-frames, and all P-frames of a GOP (up to 59) for up to eight GOPs. We use the full official split (1,200
training, 100 validation and 670 test clips) and keep the supervision budget of the VATEX study: one of about
40 captions per clip and epoch for 10 epochs (12,000 caption samples). Table~\ref{tab:msvd} reports three seeds.
The one-token readout beats the CoCap-style tokens on every seed ($+19.8$, $+12.9$, $+10.1$; pooled $+14.2$, CI
$[+10.3, +18.1]$) and the frozen context tokens by 10.1 (CI $[+6.4, +14.0]$). In this experiment, adding action tokens to the context tokens lowers CIDEr on every seed ($-4.8$, $-3.9$,
$-3.6$; pooled $-4.1$, CI $[-7.5, -0.7]$), but repeated runs (below) show that differences of this size lie
within run-to-run variation. With keyframes twice as sparse as in our
VATEX encoding and 3.7$\times$ more P-frames per GOP, motion still does not help. The warm-started row
transfers VATEX weights trained with 16 P-frames per GOP, and the action encoder's position table for 59
P-frames had to be re-initialized, so it measures cross-dataset transfer rather than better-trained motion.

\paragraph{Clean motion vectors.}
x264's default lets each block of a P-frame reference one of up to three earlier frames, and the cache does
not record which one, so motion vectors to different references are mixed. To remove this confound we
re-encode MSVD with a single reference frame, otherwise unchanged, and retrain the CoCap-style and
context-only configurations (Table~\ref{tab:msvd}, bottom). The I-frames are unaffected (cached features
identical, cosine 1.000). The CoCap-style tokens score as before (95.28 vs.\ 94.91; $+0.4$, CI $[-1.9, +2.7]$),
so the single-reference encoding gives no statistically resolved improvement in the tested motion branch. The context-only runs repeat the first experiment
with bit-identical inputs and the same configuration, yet score 92.22, 99.10 and 90.42, up to 8.4 CIDEr from
the first runs at the same seed. Their training losses match bit for bit for seven optimizer steps, so data
order, caption sampling and initialization agree, and then drift apart; validation CIDEr differs from the
first epoch on, not only at the selected checkpoint. Two repeated runs with deterministic GPU kernels match
bit for bit over all 93 steps we ran, whereas two default runs again part after seven, so the variation comes
from non-deterministic GPU kernels, amplified by training (supplementary Sec.~S3). With single-reference
vectors, action tokens change CIDEr by $+1.4$ (CI $[-1.6, +4.2]$) against these runs; with the default encoding
the change was $-4.1$. Neither is a reliable effect of motion. The readout's lead is far larger: it beats the
single-reference CoCap-style tokens by 13.9 CIDEr (CI $[+9.9, +18.1]$), and every readout run beats every
other MSVD run.

\begin{table}[t]
\centering
\small
\caption{MSVD (full official split) in CoCap's own encoding: test CIDEr on 670 clips, three seeds.
Top: x264's default of up to three reference frames; bottom: one reference frame.
$^\dagger$Motion weights from VATEX in our encoding. $^\ddagger$Same input as the context-only row of the top
block, so the difference is run-to-run variation from non-deterministic GPU kernels (Sec.~\ref{sec:msvd}).}
\label{tab:msvd}
\setlength{\tabcolsep}{3.5pt}
\begin{tabular}{@{}l c c c c@{}}
\toprule
 & \multicolumn{3}{c}{Seed} & \\
\cmidrule(lr){2-4}
Decoder input & 42 & 43 & 44 & Mean $\pm$ std \\
\midrule
\multicolumn{5}{@{}l}{\textit{Default: up to three reference frames}} \\
CoCap-style, scratch & 94.15 & 95.34 & 95.24 & 94.91 $\pm$ 0.66 \\
CoCap-style, warm$^\dagger$ & 90.13 & 90.31 & 96.59 & 92.35 $\pm$ 3.68 \\
Context tokens only & 98.97 & 99.23 & 98.83 & 99.01 $\pm$ 0.20 \\
One-query readout & \textbf{113.92} & \textbf{108.19} & \textbf{105.31} & \textbf{109.14 $\pm$ 4.38} \\
\midrule
\multicolumn{5}{@{}l}{\textit{Single reference frame (clean motion vectors)}} \\
CoCap-style, scratch & 94.33 & 95.50 & 96.01 & 95.28 $\pm$ 0.86 \\
Context tokens only$^\ddagger$ & 92.22 & 99.10 & 90.42 & 93.91 $\pm$ 4.58 \\
\bottomrule
\end{tabular}
\end{table}

\subsection{Reading the Keyframes}\label{sec:readout}

\paragraph{The readout.}
Fine-tuning SigLIP~2's attention-pooling head has the largest effect. In the per-I-frame form, whose frozen
output is the context token, it adds 6.23 CIDEr over the frozen head (CI $[+4.6, +7.9]$), or 5.38 averaged over
three seeds. A pretrained start beats a random start in every form, by \mbox{4.2--5.5} CIDEr with intervals well
above zero (Table~\ref{tab:readout}). A head that starts at random and trains is no better than the frozen
pretrained head (29.95 vs.\ 29.21), so the gain needs both the pretrained start and fine-tuning. Extra capacity
does not explain it either: applying the trained head to ToMe cluster means scores 31.10, only 0.7 above plain
ToMe-16 (30.39). Reusing a pretrained pooler on video tokens follows VideoCoCa~\cite{yan2022videococa}, which
did not compare against a random start.

\begin{table}[t]
\centering
\small
\caption{SigLIP~2's attention-pooling head as readout: frozen, trained from a random start, or fine-tuned
from its pretrained weights (test CIDEr).}
\label{tab:readout}
\begin{tabular}{@{}l c c c c@{}}
\toprule
Readout over & Tok. & Frozen & Random & Pretrained \\
\midrule
Each I-frame & 8 & 29.21 & 29.95 & 35.44 \\
All, 1 query & 1 & -- & 29.97 & 34.13 \\
All, 16 queries & 16 & -- & 30.73 & 36.14 \\
ToMe clusters & 16 & \textbf{30.48} & \textbf{32.77} & \textbf{37.25} \\
\bottomrule
\end{tabular}
\end{table}

\paragraph{Keyframes, not single-frame bias.}
With the one-query readout, test CIDEr rises from 28.66 with the middle I-frame to 29.49, 33.64 and 34.13 with
two, four and eight I-frames ($+5.47$, CI $[+3.5, +7.4]$). The clips are therefore not single-frame
problems~\cite{lei2023singleframe,buch2022atp}: more keyframes add information that the motion branch, trained
from scratch, does not. A single middle I-frame read by one fine-tuned token already matches the CoCap-style
tokens built from eight I-frames, motion vectors and residuals (both 28.66). This coincidence is also why two
different comparisons in this paper are both $+5.47$.

\paragraph{Merging versus pooling.}
At 392 tokens ToMe beats average pooling by 14.51 CIDEr (CI $[+12.5, +16.6]$), partly because pooling lands
in the high-loss regime; with GPT-2 still frozen after epoch one, ToMe already scores 9.4 validation CIDEr
against 3.7.

\begin{figure*}[t]
\centering
\input{figures/qualitative}
\caption{Random held-out test clips. Left: four of the eight I-frames the models read. Right: a reference
caption and the captions generated with CoCap-style tokens (16), ToMe (392 tokens) and the fine-tuned
one-token readout, reproduced verbatim.}
\label{fig:qualitative}
\end{figure*}

\subsection{Token Budget and Training Length}\label{sec:budget}

One fine-tuned readout query over all 1,568 patches beats the 16 CoCap-style tokens on every seed, by 5.47
(CI $[+3.6, +7.3]$), 13.33 and 6.75 CIDEr, and its own score barely moves between seeds ($33.82 \pm 0.48$). The
token count itself matters little. Averaged over seeds, one token matches the eight per-I-frame readout tokens
(33.82 vs.\ 33.78), so the gain over the CoCap-style tokens comes from fine-tuning the head. Against ToMe-392
the one-token readout trails by 2.63 and 1.45 CIDEr on seeds 42 and 43, and leads by 7.92 on seed 44, where
ToMe-392 fell into the high-loss regime. Averaged over the three seeds, the readout leads by 1.28 CIDEr; the clip-bootstrap interval, $[+0.09, +2.51]$,
just excludes zero, but it covers clip sampling only, and with per-seed differences of opposite sign the
ordering is not established across training runs. Sixteen queries beat one on all three seeds, by about 2 CIDEr. Yet free queries converge to
near-identical tokens (mean pairwise cosine 0.90 with seed 42, and a single repeated vector with 8 or 32
queries and with seed 43), so the gain does not come from distinct content.

Five epochs leave every learning curve rising. Trained for 12 epochs, the four key configurations improve by
\mbox{2.5--6.3} test CIDEr (CoCap-style 28.66 to 31.64, context only 29.21 to 33.05, ToMe-392 36.75 to 39.23, one-token
readout 34.13 to 40.44). Their order holds, except that the one-token readout gains the most
and moves from second to first. It then leads the CoCap-style tokens by 8.80 CIDEr (CI $[+6.8, +11.0]$) and leads ToMe-392 by 1.21 (CI $[-0.8, +3.2]$, one seed).

\begin{table}[t]
\centering
\small
\caption{Per-clip latency (ms, median). Bitstream: one CPU thread, 100 test clips. Networks: SigLIP~2 on
eight I-frames, motion modules, readout and GPT-2 (20 tokens) at batch 1 on a 4\,GB laptop GPU.
Sum of the two medians; lowest per column in bold.}
\label{tab:latency}
\begin{tabular}{@{}l c c c c@{}}
\toprule
Decoder input & Tok. & Bitstream & Networks & Sum \\
\midrule
CoCap-style & 16 & 114.3 & 217.1 & 331.4 \\
ToMe & 392 & \textbf{7.3} & 217.1 & 224.4 \\
Per-I-frame readout & 8 & \textbf{7.3} & 205.0 & 212.3 \\
One-query readout & 1 & \textbf{7.3} & \textbf{203.9} & \textbf{211.2} \\
\bottomrule
\end{tabular}
\end{table}

\subsection{Training Regimes of GPT-2}\label{sec:regimes}

Fine-tuning GPT-2 on this data ends in one of two regimes, judged by the final training loss (label-smoothed
cross-entropy summed over the caption tokens). In the low-loss regime the loss is about 41 and \mbox{4--16\%} of
captions loop. In the high-loss regime it is about 52 and \mbox{30--51\%} loop. Without the first-epoch decoder
freeze, the regime flipped with the seed for ToMe and average pooling alike and moved
ToMe-392 by 5.9 validation CIDEr. With the freeze, the one-query readout stays in the low-loss regime for all
three seeds. The CoCap-style configuration, however, fell into the high-loss regime with seed 43 (50.8\%
loops, test 20.72 vs.\ 28.66), and ToMe-392 did so with seed 44 (about 40\% loops, test 25.35). With
warm-started motion encoders, the CoCap-style tokens stayed in the low-loss regime on every seed. A single-seed
comparison that involves such a configuration can be off by the full width of the regime gap.

\subsection{Efficiency}\label{sec:efficiency}

Table~\ref{tab:latency} separates reading the bitstream from running the networks. On the networks alone, the
one-token readout saves only 6\% (203.9 vs.\ 217.1\,ms), far less than its 16-fold token reduction, because
GPT-2's 20-step decoding (about 8\,ms per token) dominates. The larger difference lies in the bitstream.
Decoding only the I-frames takes 7.3\,ms per clip, decoding every frame takes 44.1\,ms, and CoCap's reader,
which returns every frame with its motion vectors and residuals, takes 114.3\,ms. In both readers we tested, motion vectors and residuals are available only once every frame is decoded.
Against the faster one (44.1\,ms) the keyframe-only pipeline needs 19\% less time per clip, and against
CoCap's reader 36\% less. These totals add per-stage medians; we did not time complete pipelines end to end.

\subsection{Qualitative Results}\label{sec:qualitative}

Figure~\ref{fig:qualitative} shows three test clips drawn at random with a fixed seed, not selected by hand.
All configurations name the main actor but often miss the specific object or action; only the one-token
readout sees that the third clip shows someone riding a bike.

\section{Discussion}\label{sec:discussion}

\paragraph{Scope of the motion result.}
The result is conditional. When the motion encoders are trained from scratch on 12,500 caption samples, next
to an I-frame encoder with large-scale pretraining, the CoCap-style motion branch adds nothing. When they are
warm-started from a model trained on $48\times$ the caption samples, it adds 1.3 CIDEr on frozen context tokens.
That gain is not significant across seeds but agrees with CoCap's full-data ablation~\cite{shen2023cocap}
and with motion encoders pretrained at scale~\cite{cope2026}. On MSVD in CoCap's own encoding, with sparser keyframes, all P-frames of a GOP and even clean single-reference
motion vectors, action tokens bring no reliable gain. Less expected is that better-trained motion did not complement
a fine-tuned readout: next to fine-tuned context tokens, warm-started action tokens lowered VATEX CIDEr by 4.2.
Under limited supervision the practical order is clear. Read the keyframes well first,
since a fine-tuned readout ($+5.4$) and more keyframes ($+5.5$) each move test CIDEr far more than motion
does. A motion branch that complements a fine-tuned readout, instead of competing with it, remains an open
problem.

\paragraph{Limitations.}
Training uses 2,500 VATEX clips with one caption per clip and epoch, so absolute CIDEr is far below published
full-data numbers, and the conclusions may change with more data: we test one training-set size per dataset, so how the value
of motion grows with caption data is open beyond the warm-start control. We test one decoder
(GPT-2 small), one encoder family (SigLIP~2 with a frozen backbone, whereas CoCap fine-tunes its I-frame
encoder) and two datasets, with no drone or surveillance footage and no timing on an edge device. VATEX clips
have an I-frame every second and MSVD clips every two seconds; still sparser keyframes, common in streaming
and aerial video, may favor motion. The MSVD warm start transfers VATEX weights across datasets and
encodings. The single-reference control was run on MSVD only; on VATEX the motion vectors mix up to three reference
frames. Non-deterministic GPU kernels make repeated MSVD runs at the same seed differ by up to 8.4 CIDEr, more
than the action-token effects we measure there. We did not
study motion encoders pretrained for alignment, as in CoPE-VideoLM~\cite{cope2026}. Most configurations use one seed, while the nine behind the main and motion claims and the six MSVD configurations use three. Bootstrap intervals cover clip sampling
only.

\section{Conclusion}\label{sec:conclusion}

Under limited supervision, how a pretrained caption decoder reads the keyframes of compressed video matters
more than codec motion. A fine-tuned attention-pooling readout, where one token suffices, beats the 16 CoCap-style tokens on every
seed of VATEX and of MSVD in CoCap's own encoding, and needs only I-frames. A positive mean motion gain appeared only with warm-started encoders on frozen context tokens, and it was not
significant across seeds. For devices that receive H.264 streams, the keyframe readout is the first
component to get right, and motion the second.

{
    \small
    \bibliographystyle{ieeenat_fullname}
    \bibliography{references}
}

\appendix
\newcommand{\mainref}[2]{\ref{#1}}
\setcounter{section}{0}\setcounter{table}{0}\setcounter{figure}{0}
\renewcommand{\thesection}{S\arabic{section}}\renewcommand{\theHsection}{S\arabic{section}}
\renewcommand{\thetable}{S\arabic{table}}\renewcommand{\theHtable}{S\arabic{table}}
\renewcommand{\thefigure}{S\arabic{figure}}\renewcommand{\theHfigure}{S\arabic{figure}}
\twocolumn[{\centering\Large\bfseries Supplementary Material\par\vspace{1.5em}}]
\section{All Configurations}\label{sec:s-all}
Table~\ref{tab:s-all} lists the 34 VATEX configurations of the main study (seed 42, protocol of Sec.~\mainref{sec:protocol}{3.3} of the
main paper), grouped by family.

\begin{table*}[t]
\centering
\small
\caption{All 34 configurations (seed 42), grouped by family and sorted by held-out test CIDEr within each
group, with validation CIDEr on the 500 development clips. Loop \% (validation): captions that repeat a word
trigram (\mbox{4--16\%} in the low-loss training regime, 30\% or more in the high-loss one). Token cos.: mean
pairwise cosine of the decoder's visual tokens (1.00: all identical). Single seed: the same configuration can
vary by \mbox{1--8} CIDEr across seeds (main paper, Tables 1 and 3), so small gaps are indicative.}
\label{tab:s-all}
\begin{tabular}{@{}l c c c c c c c c@{}}
\toprule
 & & \multicolumn{2}{c}{CIDEr} & \multicolumn{3}{c}{Test} & & \\
\cmidrule(lr){3-4}\cmidrule(lr){5-7}
Configuration & Tokens & Val & Test & BLEU-4 & METEOR & ROUGE-L & Loop \% & Token cos. \\
\midrule
\csname @@input\endcsname figures/all_configs.tex
\bottomrule
\end{tabular}
\end{table*}

\section{Training Length and Training Regimes}\label{sec:s-regimes}

\begin{table}[t]
\centering
\small
\caption{Five versus twelve training epochs on VATEX (seed 42, CIDEr).}
\label{tab:s-longer}
\begin{tabular}{@{}l c c c c@{}}
\toprule
 & \multicolumn{2}{c}{5 epochs} & \multicolumn{2}{c}{12 epochs} \\
\cmidrule(lr){2-3}\cmidrule(lr){4-5}
Decoder input & Val & Test & Val & Test \\
\midrule
CoCap-style & 37.30 & 28.66 & 41.83 & 31.64 \\
Context only & 37.87 & 29.21 & 42.80 & 33.05 \\
ToMe-392 & \textbf{47.30} & \textbf{36.75} & 52.45 & 39.23 \\
Readout, 1 token & 45.28 & 34.13 & \textbf{52.57} & \textbf{40.44} \\
\bottomrule
\end{tabular}
\end{table}

\begin{figure}[t]
\centering
\begin{tikzpicture}
\begin{axis}[
  width=\columnwidth, height=5.2cm,
  xmin=0, xmax=1570, ymin=35, ymax=120,
  xlabel={Optimizer step}, ylabel={Training loss (smoothed)},
  ymajorgrids, major grid style={black!12, line width=0.4pt},
  axis line style={black!45}, tick style={black!45}, tick label style={font=\scriptsize},
  label style={font=\footnotesize}, scaled x ticks=false,
  legend style={font=\scriptsize, draw=none, fill=white, fill opacity=0.85, text opacity=1,
                at={(0.98,0.97)}, anchor=north east, legend cell align=left, row sep=-1pt},
]
\addplot[cTome, line width=1pt] table[x=step, y=tome25_seed42] {figures/data/regimes.dat};
\addlegendentry{ToMe-392, seed 42 (CIDEr 43.5)}
\addplot[cTome, line width=1pt, densely dashed] table[x=step, y=tome25_seed43] {figures/data/regimes.dat};
\addlegendentry{ToMe-392, seed 43 (CIDEr 37.6)}
\addplot[cPool, line width=1pt] table[x=step, y=pool25_seed42] {figures/data/regimes.dat};
\addlegendentry{Avg.\ pooling-392, seed 42 (CIDEr 35.4)}
\addplot[cPool, line width=1pt, densely dashed] table[x=step, y=pool25_seed43] {figures/data/regimes.dat};
\addlegendentry{Avg.\ pooling-392, seed 43 (CIDEr 34.8)}
\end{axis}
\end{tikzpicture}
\caption{Two training regimes without the first-epoch decoder freeze (validation CIDEr in the legend). Runs split within the first
hundred optimizer steps into a low-loss regime (about 41) and a high-loss regime (about 52, with about 30\%
looping captions). The regime flipped with the seed: ToMe lands low with seed 42 and high with seed 43, average pooling the
other way round.}
\label{fig:regimes}
\end{figure}
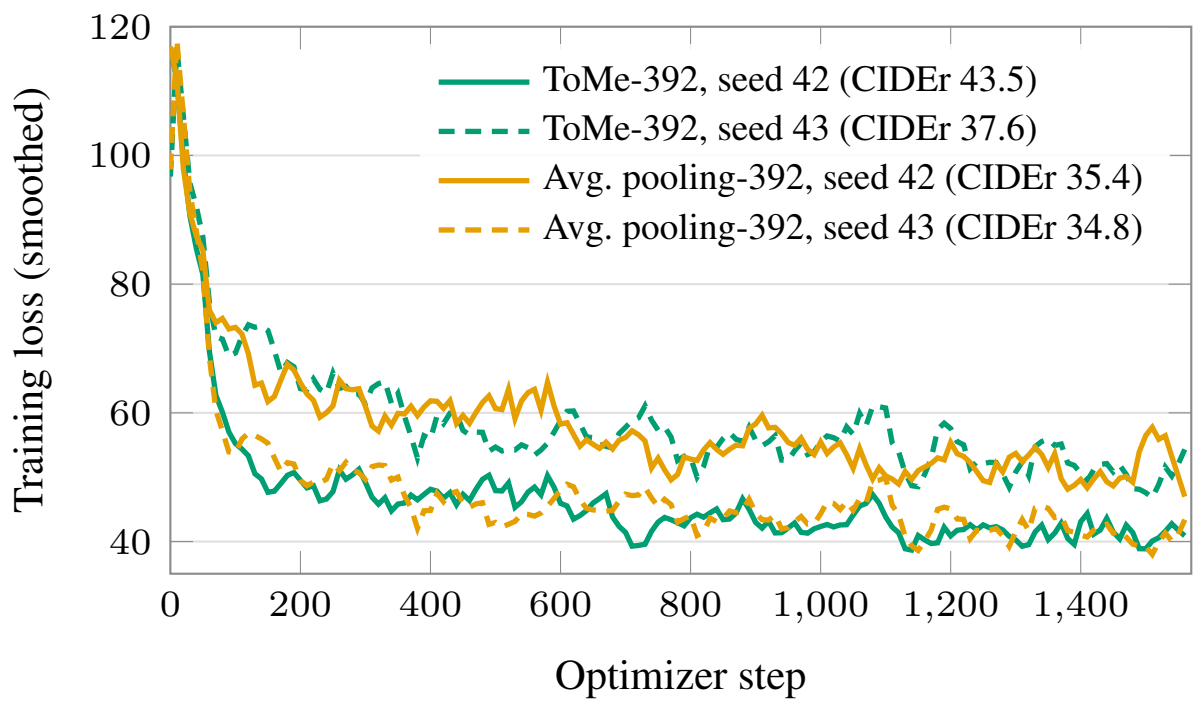

Table~\ref{tab:s-longer} gives the 12-epoch check of Sec.~\mainref{sec:budget}{4.6}. Figure~\ref{fig:regimes} shows the two training
regimes of the GPT-2 decoder discussed in Sec.~\mainref{sec:regimes}{4.7} for runs without the first-epoch decoder freeze.

\section{Reproducibility}\label{sec:s-repro}

\paragraph{Clips.}
VATEX: an 8,000-clip selection of the public VATEX splits (5,000 train, 1,000 validation, 2,000 test). The
study uses 2,500 of the 5,000 training clips (drawn with seed 42), 500 of the 1,000 validation clips and
1,000 of the 2,000 test clips (both drawn with seed 4242); training uses one of the ten English captions per clip and epoch. The
three ID lists are included with this supplementary material. MSVD: the official split (1,200 / 100 / 670
clips, 48,774 / 4,290 / 27,763 captions) as distributed with the clips in a public mirror of the dataset;
one caption per clip and epoch, 10 epochs.

\paragraph{Video encoding.}
VATEX (main study): H.264 (x264) at $224\times224$, \texttt{keyint=30}, no B-frames, x264 defaults otherwise
(3 reference frames). MSVD (CoCap's encoding): \texttt{ffmpeg -i in -an -vf "scale=<short side 240>"
-c:v libx264 -pix\_fmt yuv420p -x264-params keyint=60:min-keyint=60:scenecut=0:bframes=0 out.mp4}; the
single-reference control adds \texttt{:ref=1}.

\paragraph{Feature cache.}
One file per clip: SigLIP~2 base/16 (\texttt{google/siglip2-base-patch16-224}, frozen) last-layer patch tokens
and pooled embedding of each I-frame ($224\times224$ center crop), stored in bf16; L0 motion vectors read with
CoCap's reader on the codec's quarter-resolution grid, center-cropped to $56\times56$; residuals center-cropped to
$224\times224$ and area-reduced $4\times4$ to $56\times56$. Up to eight GOPs are sampled uniformly; clips with fewer
GOPs are padded and masked. VATEX uses 16 of the up to 29 P-frames per GOP (random in training, uniform in
evaluation); MSVD uses all P-frames (up to 59).

\paragraph{Models and readouts.}
GPT-2 small prefix decoder; the tied token embedding is frozen, the twelve blocks train. Visual tokens pass a
LayerNorm and a linear layer and carry a Fourier embedding of their $(t,y,x)$ source coordinates (size-weighted
means for merged tokens) through a zero-initialized projection; captions start with a BOS token. Readouts use
SigLIP~2's multihead attention-pooling head (probe query, multihead attention, LayerNorm, MLP) on a subset of
patch tokens: one I-frame, one ToMe cluster, or all patches with $K$ queries initialized at the probe plus
Gaussian noise of 0.02 times the probe's standard deviation. ToMe uses bipartite matching with size-weighted
averaging and tracks which original patches each merged token owns.

\paragraph{Training.}
BertAdam (weight decay 0.01 on weight matrices, 0 on biases and norms), learning rate $10^{-4}$ for new
modules and $5\times10^{-5}$ for GPT-2, 10\% linear warm-up, then decay by 0.8 per epoch; label-smoothed
cross-entropy (0.1) summed over caption tokens; captions up to 32 tokens; physical batch 1 with 8-step
gradient accumulation; bf16 mixed precision; no gradient clipping; GPT-2 frozen for the first epoch.

\paragraph{Warm start.}
A CoCap model (the same SigLIP~2 features, CoCap's motion, residual and action encoders and its 2-layer
multimodal decoder) trained end to end on all ten captions of the 5,000 VATEX training clips of the selection
for 12 epochs, VATEX encoding with 16 P-frames per GOP; the checkpoint with the best validation CIDEr (epoch 12)
was used. Its 88 motion, residual and action encoder tensors initialize the warm-started runs and train at
0.25 times the learning rate. On MSVD (59 P-frames per GOP) the action encoder's temporal position table cannot
be transferred and is re-initialized (87 of 88 tensors).

\paragraph{Evaluation and statistics.}
Greedy decoding; CIDEr-D, BLEU-4, METEOR and ROUGE-L from the COCO caption toolkit with PTB tokenization;
test scores use the checkpoint with the best validation CIDEr. Paired comparisons use per-clip CIDEr-D
differences between two runs on the same test clips, with 5,000 bootstrap resamples of the clips and the
2.5th and 97.5th percentiles. Pooled comparisons average each clip's difference over the seeds before the
bootstrap. Seed-level intervals are Student-$t$ intervals over the per-seed mean differences.

\paragraph{Run-to-run variation.}
Seeding (PyTorch Lightning's \texttt{seed\_everything}, data-loader workers included) fixes initialization, data
order and caption sampling, but training uses PyTorch's default CUDA kernels, which are not deterministic. To
find the source of the MSVD differences in Sec.~\mainref{sec:msvd}{4.4} of the main paper we compared per-step training losses.
(i)~The two seed-44 context-only runs of Table~\mainref{tab:msvd}{4} have bit-identical I-frame features in all 1,970 clips and
Hydra configurations that differ only in the cache path; their losses agree bit for bit for the first seven
optimizer steps and then diverge. Their validation CIDEr differs from the first epoch on (21 vs.\ 0 after
epoch 1, 127 vs.\ 116 after epoch 10), and checkpoint selection picks epochs 10 and 8 (test 98.83 vs.\ 90.42).
(ii)~Two new default runs of the same configuration, each stopped after 90 seconds (about 100 optimizer
steps), again agree for seven steps and then diverge, by up to 43\% in loss. (iii)~Two runs with
deterministic kernels (Lightning \texttt{deterministic=True}, \texttt{CUBLAS\_WORKSPACE\_CONFIG=:4096:8}) agree
bit for bit over all 93 common steps. The variation therefore comes from non-deterministic GPU kernels,
amplified by training; seeds fix everything else.

\paragraph{Compute.}
All reported runs were trained on one laptop RTX 3050 (4 GB) in bf16; a 5-epoch VATEX run takes 0.5--1.5 hours
and a 10-epoch MSVD run 0.4--0.8 hours. Software: PyTorch 2.5.1, PyTorch Lightning 2.5.1, Transformers 4.57.6.

\end{document}